\documentclass[12pt,a4paper]{article}

\usepackage[cmex10,intlimits]{amsmath}
\usepackage{color}
\usepackage{a4wide}
\usepackage{rotating}
\usepackage{cite}
\usepackage{tikz}
\usepackage{bm}
\usepackage{tipa}
\usepackage[utf8]{inputenc}
\usepackage[T1]{fontenc}

\usepackage{hyperref}       
\usepackage{url}            
\usepackage{booktabs}       
\usepackage{amsfonts}       
\usepackage{nicefrac}       
\usepackage{microtype}      
\usepackage[intlimits]{amsmath}
\usepackage{amssymb}
\usepackage{mathptmx}

\usepackage{subcaption}
\usepackage{graphicx}

\makeatletter
\def\blfootnote{\xdef\@thefnmark{}\@footnotetext}
\makeatother

\title{Nearest Neighbor-based Importance Weighting\footnote{Presented at MLSP 2012,  Santander, Spain.}}
\author{Marco~Loog$^{1,2}$ \medskip \small \\
\small
\begin{tabular}{c}
$^1$University of Copenhagen, Denmark \\
$^2$Delft University of Technology, The Netherlands
\end{tabular}}

\begin{document}

\maketitle

\begin{abstract}
Importance weighting is widely applicable in machine learning in general and in techniques dealing with data covariate shift problems in particular. A novel, direct approach to determine such importance weighting is presented. It relies on a nearest neighbor classification scheme and is relatively straightforward to implement. Comparative experiments on various classification tasks demonstrate the effectiveness of our so-called nearest neighbor weighting (NNeW) scheme.  Considering its performance, our procedure can act as a simple and effective baseline method for importance weighting. 
\end{abstract}

\newpage


\section{Introduction}\label{sect:intro}

A standard assumption in many approaches to supervised learning is that the source distribution $P_\varsigma$ of the samples that are used for training the learner is the same as the target distribution $P_\tau$ encountered by the learner in its operational phase.  In other words, it is assumed that the training instances are representative of the new, as yet unseen test instances.  This assumption is often violated, however, and various techniques have been studied to deal with the situation in which there is some mismatch between source and target.  The settings considered range from so-called data set shift to transfer learning and variations on these themes \cite{pan10a,quionero09a}.

In this section, we first introduce the problem of learning under covariate shift and a specific solution to it through importance weighting. Subsection \ref{sect:overv} provides an overview of relevant and related work.  Subsection \ref{sect:contr} then sketches the main contribution of the current work and acts as an overture to the remainder of the paper.

\subsection{Covariate Shift and Importance Weighting}

In the extreme case in which the source and target distributions have absolutely nothing in common and no prior knowledge about a possible link is available, one generally is not in the position to learn anything about the target from the source.  The particular presupposition we make here and which relates source and target is that the conditional probabilities of the output values $y$ given the input vector $x$ are equal under the two regimes, i.e.,
\begin{equation}
P_\varsigma(y|x) = P_\tau(y|x) \, .
\end{equation}
Given this condition, $P_\varsigma(x)$ and $P_\tau(x)$ can still be almost arbitrarily different from each other. The setting is known as learning under covariate shift \cite{shimodaira00a} and can be considered a specific form of the more general learning under sample selection bias \cite{zadrozny04a} (cf.\ \cite{srinivasan02a}).

It can be easily demonstrated that in this situation the expected loss of the source and the target can be related to each other in the following way \cite{huang07a,pan10a,quionero09a,shimodaira00a,sugiyama08b,zadrozny04a}:
\begin{equation}
\mathsf{E}_{P_\varsigma}[w(x) \ell(x,y,\theta)] = \mathsf{E}_{P_\tau}[ \ell(x,y,\theta)] \, ,
\end{equation}
with
\begin{equation}\label{eq:ratio}
w(x) := \frac{P_\tau(x)}{P_\varsigma(x)} \, .
\end{equation}
In the foregoing, $\ell(x,y,\theta)$ is the loss for the observation $(x,y)$ under model parameters $\theta$.  The crucial ingredient is the so-called importance weighting function $w(x)$ that only depends on the input variables and their marginal distributions, reweighing the source data in such a way that the expected loss for the target data can be determine through knowledge about the source distribution.

In a realistic setting of learning under covariate shift, we are given a finite sample $\{(x_i,y_j)\}_{i=1}^{N_\varsigma}$ from the source distribution $P_\varsigma$ and a finite sample $\{\xi_i\}_{i=1}^{N_\tau}$ from the input target $P_\tau$. The currently dominant approach to approximate the minimization of the loss in the target domain based on the source data consists of two steps:
\begin{enumerate}

\item

determine an estimate $\hat{w}$ of $w$ for all training points $x_i$ using the source data $x_i$ and targets $\xi_i$;

\item

minimize the empirical loss given by
\begin{equation}
\sum_{i=1}^N \hat{w}(x_i) \ell(x_i,y_i,\theta)
\end{equation}
with respect to $\theta$.

\end{enumerate}

\subsection{Estimating the Importance Weights}\label{sect:overv}

A straightforward approach to estimating the importance weights is to perform a density estimation of the input data for $P_\varsigma$ and $P_\tau$ separately, e.g.\ by means of a Parzen estimator, and calculate $w$ as the ratio of both these estimates in accordance with Equation \ref{eq:ratio}.  It has been argued and empirically demonstrated, however, that approaches that estimate $w$ directly are typically superior to the former approach \cite{huang07a,sugiyama08a,sugiyama08b}.  Especially the so-called Kullback-Leibler importance estimation procedure (KLIEP) has shown to perform very well in the two-step procedure to covariate shift, both in classification and regression tasks \cite{sugiyama08a,sugiyama08b,tsuboi09a}.

KLIEP determines an estimate of $w$ which minimizes the Kullback-Leibler divergence $KL(P_\tau||w P_\varsigma)$ between $w P_\varsigma$ and $P_\tau$.  The weighting function $w$ is typically modeled as a nonparametric regression function employing a certain number of Gaussian-kernels located at target samples $\xi_i$.  The objective function for KLIEP is convex in this case and can in principle be solved easily.  Still, the optimization might be quite time consuming. For that reason Sugiyama and collaborators proposed a method that is computationally more efficient and they formulate the estimation of the weighting function as a least-squares function fitting problem \cite{kanamori08a,kanamori09a}. The particular version that seems to perform rather well is coined unconstrained least-squares importance fitting (uLSIF), which allows a closed-form solution.  A related importance weighting approach, sharing some of the pros and cons of KLIEP and uLSIF, is kernel mean matching \cite{huang07a}.

Another approach to importance weighting uses the neat idea that any standard probabilistic classifier, built to separate source data from target data, estimates the ratio in Equation (\ref{eq:ratio}) either explicitly or implicitly.  This approach has been studied especially in combination with discriminative classifiers such as logistic regression, which like KLIEP and uLSIF avoids estimating the underlying data densities explicitly \cite{bickel07a,bickel09a} (cf.\ \cite{qin98a}). We should point out, that the authors of \cite{bickel07a,bickel09a} in fact aim at an integrated approach to estimating the model for target data, avoiding the two-step method sketched above, which they deem ad hoc.

We refer the reader to \cite{sugiyama10a} for a more extensive and comprehensive overview of density ratio estimators.

\subsection{Contribution}\label{sect:contr}

In this work, we introduce yet another approach to determine importance weights for bias correction in covariate shift.  The approach is very easy to implement: all one needs is a 1-nearest neighbor classifier.  The efficiency of the approach is basically determined by the efficiency of the implementation of this nearest neighbor classifier, but even in the case where one relies on a brute-force search for nearest neighbors, the computational complexity may generally not be the bottleneck.

Moreover, the approach proposed is very stable. There is no optimization problem to solve as is the case for KLIEP and no operations like matrix inversions, as for instance used in uLSIF, are needed.  As a matter of fact, without claiming that these problem cannot be alleviated, we do already want to mention that experiments we carried out with an implementation of KLIEP from the original author\footnote{\url{http://sugiyama-www.cs.titech.ac.jp/~sugi/software/KLIEP/}} showed that the procedure may behave unsatisfactorily on various classification problems from the UCI Machine Learning Repository \cite{frank10a}.  It sometimes does not attain a well-behaved solution for the weighting function in those cases.  Also uLSIF\footnote{\url{http://sugiyama-www.cs.titech.ac.jp/~sugi/software/uLSIF/}} fails occasionally on some data set.  We will elaborate on these results in Subsection \ref{sect:res}. All in all, the experimental results in Section \ref{sect:exp} demonstrate convincingly that our easy-to-implement importance weighting scheme can act as a simple and effective baseline method complementary to other approaches.  First, however, the next section presents our new method, which we name nearest neighbor weighting (NNeW).

\section{Nearest Neighbor Weighting}

To come to an estimate of the weighting function $w$, we cannot simply rely on the empirical distributions of the source and target data.  If the input domains are continuous, doing so would generically result in weights zero for all training samples $(x_i,y_i)$ as the source and target input samples do typically not coincide.  To get to a nontrivial estimate we have to rely on some form of interpolation (and/or extrapolation), whether through a reproducing kernel Hilbert space approach, Parzen density estimates, or nonparametric regression.

The interpolation controls which target samples $\xi_i$ exercise influence on which source samples $x_i$ and what magnitude this influence is.  A natural choice for the region of influence $V_i$ associated with any $x_i$ is the Voronoi cell containing that $x_i$ \cite{aurenhammer91a}.  The cell $V_i$ is the convex set of those points in the feature space for which there are no source samples nearer than $x_i$.  The collection of all $N$ cells constitute a Voronoi or Dirichlet tessellation of the feature space.  Subsequently, based on this natural tessellation of space, we can define our NNeW scheme and estimate the weights of every sample $x_i$ simply by counting the number of target samples $\xi_i$ that are within its associated cell, i.e.,
\begin{equation}\label{eq:nnew}
\hat{w}(x_i) =  | V_i \cap \{\xi_j\}_{j=1}^{N_\tau} | \, ,
\end{equation}
where $|A|$ measures the cardinality of the set $A$.

Employing a Voronoi tessellation makes sure that, on the one hand, no target data is wasted as the cells cover the whole space and every $\xi_j$ contributes to the weight of some $x_i$.  On the other hand, it also avoids the estimate to be overly smooth by making the regions of influence for every $x_i$ no larger than necessary to cover the whole space.  While there are other partitions that fulfilling such criteria, a Voronoi tessellation is a natural data-dependent choice.

\subsection{Approximation to the True Weights}

With increasing numbers of source and target data, $\hat{w}$ in Equation (\ref{eq:nnew}) will be approximately proportional to the true weighting function $w$, because for every small cell $V_i$ that shrinks with increasing sample sizes, we have the following approximation:
\begin{equation}
\begin{split}
\hat{w}(x_i) & = | V_i \cap \{\xi_j\}_{j=1}^{N_\tau} | = \int_{V_i} \sum_{j=1}^{N_\tau} \delta(x - \xi_i) \, dx \\
& \approx N_\tau \int_{V_i} P_\tau(x) \, dx \approx N_\tau P_\tau(x_i) \int_{V_i} 1 \, dx \\
& \approx \frac{N_\tau P_\tau(x_i)}{N_\varsigma P_\varsigma(x_i)} \propto \frac{P_\tau(x_i)}{P_\varsigma(x_i)} \, ,
\end{split}
\end{equation}
for fixed $N_\varsigma$ and $N_\tau$. The derivation uses a property of a Voronoi tessellation, which \cite{learned-miller04a} refers to as a near uniform partition, roughly stating that the integral of the distribution $P_\varsigma$ over each cell is approximately constant, which leads to:
\begin{equation}
P_\varsigma(x_i) \int_{V_i} 1 \, dx \approx \int_{V_i} P_\varsigma(x) \, dx \approx \frac{1}{N_\varsigma} \, .
\end{equation}
This property has been argued to hold in, for instance, \cite{miller03a} and \cite{learned-miller04a} and relates to the intensity of a spatial Poisson processes as treated in \cite{moller94a,okabe00a}.

\subsection{Use of Nearest Neighbor Classifier}

All $\hat{w}(x_i)$ can be obtained by training a nearest neighbor classifier in which every source data point $x_i$ makes up its own class.  A target sample $\xi_i$ adds one to the weight  $\hat{w}(x_i)$ if it is classified to the class determined by $x_i$.  The 1-nearest neighbor classifier divides up the feature space exactly like the Voronoi tessellation corresponding to $\{x_i\}_{i=1}^{N_\varsigma}$ with every cell associated to a single class.

We remark that the use of the nearest neighbor classifier may not be really essential, as most classifiers that can deal with a single sample per class would lead to the same partitioning of feature space into $N_\varsigma$ classes.

\subsection{Laplace Smoothing}

Realizing that our method may run into problems when too many Voronoi cells are empty, we decided to also conduct experiments with what is occasionally referred to as Laplace smoothing.  In effect, this means that instead of employing Equation (\ref{eq:nnew}), every cell is assumed to have already one count and we estimate the weights to be proportional to \begin{equation}\label{eq:nnew+1}
\hat{w}(x_i) = | V_i \cap \{\xi_j\}_{j=1}^{N_\tau} | + 1 \, .
\end{equation}
We note that this will also have regularizing effect as it avoids the weights to get overenthusiastically biased towards the test set.  In the experimental section, we refer to this instantiation of our technique as NNeW+1.

\section{Experimental Setup and Results}\label{sect:exp}

We carried out experiments with fourteen classification data sets from the UCI Machine Learning Repository \cite{frank10a}.  We compared our NNeW scheme, regular and smoothed, to two of the state-of-the-art importance weighting schemes discussed in Section \ref{sect:intro}, KLIEP \cite{sugiyama08a,sugiyama08b} and uLSIF \cite{kanamori08a,kanamori09a}, and the Parzen approach sketched at the beginning of that section.  For KLIEP and uLSIF, we used the two implementations provided by one of the authors (see Footnotes 1 and 2).

\subsection{Classifiers}

In our experiments, we used two different classifiers: classical linear discriminant analysis (LDA) and quadratic discriminant analysis (QDA) \cite{mclachlan92a,ripley96a}.  Both can be adapted easily so as to take into account sample weights by replacing the estimates of the means, covariance matrices, and class priors by their weighted versions.

\subsection{UCI Data Sets and Data Processing}

\begin{table*}[th]\scriptsize
\begin{center}\caption{Basic properties of the fourteen original UCI Machine Learning Repository data sets: data set name, number of objects, feature dimensionality, number of classes, size of the smallest class, and size of the largest class.}\label{tab:data}
\begin{tabular}{l|ccccc|}
data set & \#objects & dimensionality & \#classes & smallest & largest
\\ \hline
{\tt clean2} & 6598 & 167 & 2 & 1017 & 5581 \\
{\tt haberman} & 306 & 3 & 2 & 81 & 225 \\
{\tt iris} & 150 & 4 & 3 & 50 & 50 \\
{\tt newthyroid} & 215 & 5 & 3 & 30 & 150 \\
{\tt parkinsons} & 195 & 22 & 2 & 48 & 147 \\
{\tt pima} & 768 & 8 & 2 & 268 & 500 \\
{\tt sat} & 6435 & 36 & 6 & 626 & 1533 \\
{\tt spambase} & 4601 & 57 & 2 & 1813 & 2788 \\
{\tt transfusion} & 748 & 3 & 2 & 178 & 570 \\
{\tt vehicle} & 846 & 18 & 4 & 199 & 218 \\
{\tt vowel} & 990 & 10 & 11 & 90 & 90 \\
{\tt vowel\_context} & 990 & 13 & 11 & 90 & 90 \\
{\tt wdbc} & 569 & 30 & 2 & 212 & 357 \\
{\tt wine} & 178 & 13 & 3 & 48 & 71
\end{tabular}
\end{center}
\end{table*}

We carried out experiments on fourteen UCI classification problems.  Table \ref{tab:data} provides an overview of the data sets ultimately chosen, together with some of their basic properties, namely: the number of objects in the data sets, the dimensionality of the feature space, the number of classes, and the size of the smallest and largest class.

In order to avoid potential problems with singular covariance matrices in LDA and QDA, in a first preprocessing step, we reduced the dimensionality of all data sets by PCA such that a fraction of $0.999$ of the total variance in the data is retained.  In a second step we, we process the data to model a particular covariate shift in the underlying feature distribution.  As target data, we take the data sets as they are after PCA.  The source data is biased in the following way: we consider the two-dimensional space defined by the two first principle components obtained through PCA and reduce the data in quadrants {\sc i} and {\sc iii} by a factor of approximately $5$.  The same biasing procedure in this two-dimensional subspace is applied to all data sets.  This process, in combination with the PCA that has been applied, changes the properties of all data sets. These can be found in Table \ref{tab:prepdata} and should be compared to Table \ref{tab:data}.  Note, for instance, that for some data sets the PCA has for instance no effect on the dimensionality, for others the reduction can be quite substantial as one can see for {\tt wine}.

\begin{table*}[th]\scriptsize
\begin{center}\caption{Properties of the fourteen UCI Machine Learning Repository source data sets after processing. PCA reduces the dimensionality of the original data and the biased sampling scheme (as described in the text) reduces the number of instances. Reported are the data set name, number of objects, feature dimensionality, number of classes, size of the smallest class, and size of the largest class.  Compare this table to \ref{tab:data}}\label{tab:prepdata}
\begin{tabular}{l|ccccc|}
data set & \#objects & dimensionality & \#classes & smallest & largest
\\ \hline
{\tt clean2} & 3712 & 121 & 2 & 600 & 3112 \\
{\tt haberman} & 190 & 3 & 2 & 53 & 137 \\
{\tt iris} & 95 & 4 & 3 & 28 & 35 \\
{\tt newthyroid} & 109 & 5 & 3 & 17 & 70 \\
{\tt parkinsons} & 80 & 4 & 2 & 17 & 63 \\
{\tt pima} & 469 & 6 & 2 & 146 & 323 \\
{\tt sat} & 3000 & 33 & 6 & 235 & 689 \\
{\tt spambase} & 2234 & 3 & 2 & 1085 & 1149 \\
{\tt transfusion} & 436 & 3 & 2 & 98 & 338 \\
{\tt vehicle} & 566 & 10 & 4 & 130 & 156 \\
{\tt vowel} & 579 & 10 & 11 & 26 & 74 \\
{\tt vowel\_context} & 582 & 12 & 11 & 47 & 61 \\
{\tt wdbc} & 325 & 3 & 2 & 117 & 208 \\
{\tt wine} & 121 & 2 & 3 & 31 & 58
\end{tabular}
\end{center}
\end{table*}

\subsection{Comparative Experiments}

Based on the target and the source data as described in the previous subsection, the following experiments were done for every UCI data set. Two different training sets were constructed: for one half of the source data is used and for the other a fraction of the source data is used such that (non-regularized) QDA is just trainable, i.e., in which it is made sure that the smallest class still contains the dimensionality plus one instances.  The test set is half of the target data.  Training and test set are chosen such that they do not include equal instances. The classification error is measured using both weighted LDA and weighted QDA, where the weights are obtained by NNeW, NNeW+1, KLIEP, uLSIF, and Parzen. The foregoing procedure is repeated a hundred times in which every time new, random, non-overlapping training and test sets are generated for all fourteen classification problems.

\subsection{Results and Observations}\label{sect:res}

\begin{table*}[th]\footnotesize
\begin{center}\caption{Average error rates as obtained by (weighted) LDA for the fourteen UCI data sets with weights determined by KLIEP, uLSIF, Parzen, NNeW, and smoothed NNeW.  Results using a small set of source instances, left multicolumn, and a large set of source instances, right multicolumn, are reported.  Compare to Table \ref{tab:qda}.}\label{tab:lda}
\begin{tabular}{l|ccccc|ccccc|}
& \multicolumn{5}{|c|}{dimensionality + 1} & \multicolumn{5}{|c|}{large training set} \\
data set &  KLIEP & uLSIF & Parzen & NNeW & NNeW+1 & KLIEP & uLSIF & Parzen & NNeW & NNeW+1
\\ \hline
{\tt clean2}  & --- & .176 & .154 & .091 & \bf \underline{.087} & --- & .153 & .126 & .074 & \bf \underline{.069}\\
{\tt haberman}  & \bf \underline{.314} & --- & \bf .315 & \bf .322 & \bf .320 & \bf \underline{.250} & .267 & .256 & .259 & .256\\
{\tt iris}  & \bf \underline{.049} & --- & .060 & .067 & .058 & \bf .030 & --- & .034 & .034 & \bf \underline{.029}\\
{\tt newthyroid}  & \bf .077 & .107 & \bf \underline{.076} & .088 & .082 & .074 & .099 & \bf \underline{.064} & .085 & .080\\
{\tt parkinsons}  & .245 & .355 & .252 & .239 & \bf \underline{.230} & .224 & .291 & .220 & .215 & \bf \underline{.203}\\
{\tt pima}  & --- & .344 & \bf \underline{.312} & .318 & .315 & --- & .262 & \bf .253 & .256 & \bf \underline{.251}\\
{\tt sat}  & .246 & .228 & .271 & \bf .195 & \bf \underline{.192} & .213 & .201 & .216 & .173 & \bf \underline{.169}\\
{\tt spambase}  & --- & --- & \bf \underline{.351} & \bf .361 & \bf .361 & --- & .312 & \bf \underline{.284} & .324 & .318\\
{\tt transfusion}  & \bf .284 & .294 & .298 & \bf .281 & \bf \underline{.279} & \bf \underline{.225} & .244 & .229 & .232 & .229\\
{\tt vehicle}  & .441 & .440 & .445 & \bf .429 & \bf \underline{.424} & .375 & .376 & .378 & .358 & \bf \underline{.348}\\
{\tt vowel}  & .442 & .457 & .464 & \bf \underline{.417} & .423 & .438 & .454 & .459 & \bf \underline{.412} & .422\\
{\tt vowel\_context}  & \bf .451 & .465 & .511 & .463 & \bf \underline{.450} & .422 & .422 & .476 & .427 & \bf \underline{.409}\\
{\tt wdbc}  & --- & --- & \bf \underline{.160} & .183 & .181 & --- & \bf \underline{.098} & .109 & .132 & .132\\
{\tt wine}  & .354 & .432 & \bf \underline{.327} & .345 & .340 & \bf .285 & .320 & \bf \underline{.280} & .300 & .307
\end{tabular}
\end{center}
\end{table*}

\begin{table*}[th]\footnotesize
\begin{center}\caption{Average error rates as obtained by (weighted) QDA for the fourteen UCI data sets with weights determined by KLIEP, uLSIF, Parzen, NNeW, and smoothed NNeW.  Results using a small set of source instances, left multicolumn, and a large set of source instances, right multicolumn, are reported. Compare to Table \ref{tab:lda}.}\label{tab:qda}
\begin{tabular}{l|ccccc|ccccc|}
& \multicolumn{5}{|c|}{dimensionality + 1} & \multicolumn{5}{|c|}{large training set} \\
data set &  KLIEP & uLSIF & Parzen & NNeW & NNeW+1 & KLIEP & uLSIF & Parzen & NNeW & NNeW+1
\\ \hline
{\tt clean2}  & --- & .310 & .166 & \bf \underline{.143} & .153 & --- & .461 & .148 & .080 & \bf \underline{.057}\\
{\tt haberman}  & \bf .312 & --- & \bf \underline{.309} & .363 & \bf .317 & \bf \underline{.255} & .269 & \bf .259 & .264 & .261\\
{\tt iris}  & \bf \underline{.217} & --- & .248 & .286 & \bf .226 & \bf \underline{.039} & --- & .051 & .068 & \bf \underline{.039}\\
{\tt newthyroid}  & .151 & .216 & .198 & .420 & \bf \underline{.136} & \bf .079 & .178 & .111 & .172 & \bf \underline{.076}\\
{\tt parkinsons}  & .266 & .367 & \bf \underline{.245} & .397 & \bf .247 & .253 & .285 & .252 & .254 & \bf \underline{.238}\\
{\tt pima}  & --- & \bf \underline{.348} & .350 & .368 & .352 & --- & .307 & .285 & .287 & \bf \underline{.279}\\
{\tt sat}  & .579 & .689 & .586 & .338 & \bf \underline{.328} & .324 & .341 & .348 & .194 & \bf \underline{.175}\\
{\tt spambase}  & --- & --- & \bf .405 & \bf .402 & \bf \underline{.401} & --- & .323 & \bf .309 & \bf \underline{.308} & .312\\
{\tt transfusion}  & \bf .299 & \bf .299 & \bf .299 & .353 & \bf \underline{.292} & \bf \underline{.235} & .268 & .241 & .242 & \bf \underline{.235}\\
{\tt vehicle}  & .621 & .632 & .623 & \bf \underline{.608} & \bf .615 & .371 & .378 & .370 & .324 & \bf \underline{.300}\\
{\tt vowel}  & \bf .351 & \bf .349 & .458 & .456 & \bf \underline{.348} & \bf .302 & \bf .297 & .410 & .359 & \bf \underline{.296}\\
{\tt vowel\_context}  & \bf \underline{.490} & .553 & .655 & .652 & .504 & \bf \underline{.287} & .295 & .383 & .360 & \bf .288\\
{\tt wdbc}  & --- & --- & .258 & .229 & \bf \underline{.203} & --- & .119 & \bf \underline{.072} & .078 & \bf .073\\
{\tt wine}  & .488 & .528 & \bf \underline{.401} & .428 & \bf .412 & .309 & .389 & \bf \underline{.296} & .311 & \bf \underline{.296}
\end{tabular}
\end{center}
\end{table*}

Tables \ref{tab:lda} and \ref{tab:qda} report the mean classification errors over all hundred repetitions.  Table \ref{tab:lda} shows the results for LDA both when using the small target data set to determine the covariate shift, in the multicolumn `dimensionality + 1', and when using all target data to estimate the shift, in the multicolumn `large training set'.  Table \ref{tab:qda} gives the same overview for QDA.   In underlined bold are the best averaged error rates, while according to their standard deviations of the means, the other bold faced error rates are not significantly different from this lowest error rate within the respective experiments. 

Concerning Tables \ref{tab:lda} and \ref{tab:qda}, for LDA, NNeW+1 seems to be the overall best performing approach.  For QDA this becomes even more apparent, showing NNeW+1 to be the best performing technique in sixteen out of 28 cases.  For QDA, in only five out of 28 cases it performs significantly worse than the best performing other technique, while practically it is even in these cases not very far off.  Though NNeW clearly shows worse performance than NNeW+1, it closely follows in most of the experiments, especially for LDA. Once in a while, however, its performance seems dramatically worse than NNeW+1's, e.g.\ on {\tt newthyroid}.  We expect that in these cases we see the beneficial effect of the Laplace smoothing, though we cannot say that this effect is more evident in case of the smaller training set.

uLSIF generally performs worst but this might be expected given the fact that it was developed as an approximation to KLIEP that should be able to work on large data sets as well.  Both KLIEP and Parzen seem to perform good with the former slightly better than the latter. Yet NNeW+1 more often outperforms KLIEP than the other way around. And also NNeW often improves upon the results that KLIEP attains.  On {\tt sat}, the improvements of our approach over KLIEP are most prominent, while on {\tt newthyroid}, KLIEP clearly beats NNeW.  There are, however, data sets on which uLSIF and KLIEP often simply failed to provide any proper results (indicated by dashes in both tables). uLSIF solves an unconstrained least-squares problem and to obtain a proper nonnegative solution, negative weights are simply set to zero.  This resulted regularly in too few training data points with a positive weight. In turn, this lead to singular estimates for the covariance matrices in which case LDA and QDA are not properly defined.  KLIEP occasionally suffers from an ill-behaved optimization after which no proper weights are returned.

\section{Discussion and Conclusion}

A novel approach to importance weighting for covariate shift adaptation has been presented.  It is straightforward to implement and demonstrates competitive performance in a comparative study with two state-of-the-art importance weight estimators, KLIEP and uLSIF.  Moreover, it is stable in its use as the most complicated operation necessary for its successful application is a nearest neighbor search for every target data point among the instances in the source data. Despite the, necessarily limited, experimental evaluation, we venture to conclude that our nearest neighbor weighting (NNeW), and certainly its Laplace smoothed variant: NNeW+1, may generally act as an effective and simple baseline method to determine importance weights.  NNeW+1 seems, however, generally to be preferred, not only for its performance but because, like uLSIF, also the NNeW scheme may in certain cases suffer from the fact that too many source training samples get assigned a zero weight. In the case, for example, that the target samples are confined to a small region in feature space.  For classifiers such as (weighted) LDA and QDA this is obviously problematic, but also for other classifiers this situation can be detrimental.  It is therefore certainly of interest to further investigate ways to construct smoothed versions of NNeW.

\bibliography{nnew}

\begin{thebibliography}{10}

\bibitem{pan10a}
S.J. Pan and Q.~Yang.
\newblock A survey on transfer learning.
\newblock {\em IEEE Transactions on Knowledge and Data Engineering},
  22(10):1345--1359, 2010.

\bibitem{quionero09a}
J.~{Qui\~{n}onero Candela}, M.~Sugiyama, A.~Schwaighofer, and N.D. Lawrence.
\newblock {\em Dataset shift in machine learning}.
\newblock The MIT Press, 2009.

\bibitem{shimodaira00a}
H.~Shimodaira.
\newblock Improving predictive inference under covariate shift by weighting the
  log-likelihood function.
\newblock {\em Journal of Statistical Planning and Inference}, 90(2):227--244,
  2000.

\bibitem{zadrozny04a}
B.~Zadrozny.
\newblock Learning and evaluating classifiers under sample selection bias.
\newblock In {\em Proceedings of the twenty-first International Conference on
  Machine Learning}, page 114, 2004.

\bibitem{srinivasan02a}
R.~Srinivasan.
\newblock {\em Importance sampling: Applications in communications and
  detection}.
\newblock Springer Verlag, 2002.

\bibitem{huang07a}
J.~Huang, A.J. Smola, A.~Gretton, K.M. Borgwardt, and B.~Sch\"{o}lkopf.
\newblock Correcting sample selection bias by unlabeled data.
\newblock In {\em Advances in Neural Information Processing Systems},
  volume~19, page 601, 2007.

\bibitem{sugiyama08b}
M.~Sugiyama, T.~Suzuki, S.~Nakajima, H.~Kashima, P.~von B{\"{u}}nau, and
  M.~Kawanabe.
\newblock Direct importance estimation for covariate shift adaptation.
\newblock {\em Annals of the Institute of Statistical Mathematics},
  60(4):699--746, 2008.

\bibitem{sugiyama08a}
M.~Sugiyama, S.~Nakajima, H.~Kashima, P.~von B{\"{u}}nau, and M.~Kawanabe.
\newblock Direct importance estimation with model selection and its application
  to covariate shift adaptation.
\newblock In {\em Advances in Neural Information Processing Systems},
  volume~20, pages 1433--1440, 2008.

\bibitem{tsuboi09a}
Y.~Tsuboi, H.~Kashima, S.~Hido, S.~Bickel, and M.~Sugiyama.
\newblock Direct density ratio estimation for large-scale covariate shift
  adaptation.
\newblock {\em Information and Media Technologies}, 4(2):529--546, 2009.

\bibitem{kanamori08a}
T.~Kanamori, S.~Hido, and M.~Sugiyama.
\newblock Efficient direct density ratio estimation for non-stationarity
  adaptation and outlier detection.
\newblock In {\em Advances in Neural Information Processing Systems},
  volume~20, pages 809--816, 2008.

\bibitem{kanamori09a}
T.~Kanamori, S.~Hido, and M.~Sugiyama.
\newblock A least-squares approach to direct importance estimation.
\newblock {\em Journal of Machine Learning Research}, 10:1391--1445, 2009.

\bibitem{bickel07a}
S.~Bickel, M.~Br{\"{u}}ckner, and T.~Scheffer.
\newblock Discriminative learning for differing training and test
  distributions.
\newblock In {\em Proceedings of the 24th International Conference on Machine
  Learning}, pages 81--88, 2007.

\bibitem{bickel09a}
S.~Bickel, M.~Br{\"{u}}ckner, and T.~Scheffer.
\newblock Discriminative learning under covariate shift.
\newblock {\em Journal of Machine Learning Research}, 10:2137--2155, 2009.

\bibitem{qin98a}
J.~Qin.
\newblock Inferences for case-control and semiparametric two-sample density
  ratio models.
\newblock {\em Biometrika}, 85(3):619, 1998.

\bibitem{sugiyama10a}
M.~Sugiyama, T.~Suzuki, and T.~Kanamori.
\newblock Density ratio estimation: A comprehensive review.
\newblock {\em RIMS Kokyuroku}, pages 10--31, 2010.

\bibitem{frank10a}
A.~Frank and A.~Asuncion.
\newblock {UCI} machine learning repository, 2010.

\bibitem{aurenhammer91a}
F.~Aurenhammer.
\newblock {V}oronoi diagrams -- a survey of a fundamental geometric data
  structure.
\newblock {\em ACM Computing Surveys}, 23(3):345--405, 1991.

\bibitem{learned-miller04a}
E.~Learned-Miller.
\newblock Hyperspacings and the estimation of information theoretic quantities.
\newblock Technical Report 04-104, University of Massachusetts Amherst, 2004.

\bibitem{miller03a}
E.G. Miller.
\newblock A new class of entropy estimators for multi-dimensional densities.
\newblock In {\em Proceedings of the 2003 IEEE International Conference on
  Acoustics, Speech, and Signal Processing}, volume~3, pages III--297, 2003.

\bibitem{moller94a}
J.~M{\o}ller.
\newblock {\em Lectures on Random {V}oronoi Tessellations, Volume 87 of Lecture
  Notes in Statistics}, volume~87 of {\em Lecture Notes in Statistics}.
\newblock Springer-Verlag, 1994.

\bibitem{okabe00a}
A.~Okabe, B.~Boots, K.~Sugihara, and S.N. Chiu.
\newblock {\em Spatial tessellations: Concepts and applications of {V}oronoi
  diagrams}.
\newblock Wiley Series in Probability and Mathematical Statistics, 2000.

\bibitem{mclachlan92a}
G.J. McLachlan.
\newblock {\em Discriminant analysis and statistical pattern recognition}.
\newblock Wiley \& Sons, 1992.

\bibitem{ripley96a}
B.D. Ripley.
\newblock {\em Pattern recognition and neural networks}.
\newblock Cambridge University Press, 1996.

\end{thebibliography}
\bibliographystyle{unsrt}

\end{document}